\documentclass[conference]{IEEEtran}
\IEEEoverridecommandlockouts

\usepackage{cite}
\usepackage{amsmath,amssymb,amsfonts}
\usepackage{graphicx}
\usepackage{textcomp}
\usepackage[T1]{fontenc}
\usepackage[utf8]{inputenc}
\usepackage{url}
\usepackage{microtype}
\usepackage[hidelinks,breaklinks]{hyperref}

\graphicspath{{figs/}}

\begin{document}

\title{Attention, Dynamic Wave and Hybrid\\
Sequence Mixers in Language Models}

\author{\IEEEauthorblockN{Dmytro Babarytskyi}
\IEEEauthorblockA{\textit{Kharkiv National University of Radio Electronics}\\
Kharkiv, Ukraine\\
dmytro.babarytskyi@nure.ua\\
babaritskiydmytro@gmail.com\\
ORCID: 0009-0009-3614-9601}
\and
\IEEEauthorblockN{Oleksandr Babarytskyi}
\IEEEauthorblockA{\textit{Military Law Institute}\\
\textit{Yaroslav Mudryi National Law University}\\
Kharkiv, Ukraine\\
aleksandr.babaritskiy@ukr.net\\
ORCID: 0000-0002-1690-1837}
}

\maketitle

\begin{abstract}
Large Language Models (LLMs) rely on Multi-Head Attention (MHA) for sequence
modeling, but scaling to long contexts is constrained by the quadratic cost of
attention, motivating sub-quadratic recurrent alternatives. This paper
formalizes a Dynamic Wave recurrent layer, based on input-dependent
exponential damping and cosine phase modulation, and evaluates three
10.94M-parameter configurations---Pure Transformer, Pure Wave, and Hybrid
Wave-Attention---on TinyStories under perplexity, latency, throughput and peak
VRAM profiling up to 4096 tokens. The Hybrid configuration reaches the best
perplexity, 5.95 on the full held-out split against 6.93 for the Transformer
and 7.26 for Pure Wave. The
efficiency picture, however, is governed by implementation rather than by
architecture. Against an unfused attention baseline, Pure Wave uses 78.7\%
less peak VRAM at a 4096-token context (1404.3~MB vs. 6601.6~MB); against a
fused scaled-dot-product attention kernel the same comparison yields no
advantage at all (1404.3~MB vs. 1403.6~MB), and in training mode Pure Wave is
8.3\% more expensive than the Transformer. Likewise, hoisting elementwise terms out of
the recurrence loop speeds the layer up 2.53-fold with no custom kernel,
reducing its latency gap to attention from 2.48-fold to 1.12-fold. A hybrid stack is found to
inherit the cost of its more expensive mixer rather than the advantage of its
cheaper one, its peak VRAM matching the Pure Transformer to within 0.1\%. A
closed-form memory model reproduces these measurements and predicts the context
length beyond which attention overtakes the vocabulary projection. The
findings indicate that reported efficiency gains for recurrent attention
replacements must be stated relative to a named attention backend and a named
implementation, and that hybridization is best justified on quality grounds.
\end{abstract}


% =====================================================================
\section{Introduction}\label{sec:intro}

Large Language Models (LLMs) have fundamentally transformed the paradigms of
natural language processing, computer vision, and software engineering by
enabling sophisticated text generation, contextual reasoning, and zero-shot
problem-solving \cite{radford2019,kaplan2020}. At the core of modern LLM
architectures lies the Transformer model, which relies heavily on the
Multi-Head Attention (MHA) mechanism to perform sequence modeling
\cite{vaswani2017}. This mechanism enables the model to dynamically capture
long-range global dependencies across a sequence, establishing MHA as the
standard for autoregressive language generation and deep context
understanding.

Despite its unparalleled success, scaling traditional MHA to exceptionally
long contexts remains constrained by its inherent quadratic computational
complexity $O(N^{2})$, where $N$ denotes the sequence length. As the context
window expands, the memory footprint required to store the Key-Value (KV)
cache grows linearly with context length while the cost of the attention
computation itself grows quadratically, together precipitating a critical
``memory wall'' that exhausts available Video RAM (VRAM) \cite{dao2022}. This
memory problem not only drives up computational infrastructure costs but also
leads to long-context degradation, where the model's token-processing
efficiency drops significantly, and its retrieval capabilities suffer from
structural information loss deep within lengthy prompts \cite{liu2024lost}.

To alleviate this quadratic complexity, alternative sub-quadratic sequence
mixing paradigms have been proposed, including State-Space Models (SSMs),
linear-time recurrent layers, and specialized linear attention approximations
\cite{gu2023mamba,peng2023rwkv}. Architectures such as Mamba and RWKV compress
the sequence history into a fixed-size latent state, effectively reducing
spatial complexity to $O(N)$ during inference. However, these pure recurrent
implementations frequently suffer from sequential processing latency on
parallel hardware during training and can exhibit degraded semantic
understanding. Specifically, they lack the strong global associative recall
capabilities of standard MHA, leading to suboptimal performance on tasks that
require tight, non-local token dependencies.

To address the trade-off between the expressive power of attention and the
computational efficiency of recurrent networks, recent architectural trends
have shifted toward hybrid sequence modeling paradigms
\cite{de2024griffin,lieber2024jamba}. Models like Griffin and Jamba interleave
MHA layers with gated linear recurrences or SSMs to achieve sub-quadratic
memory scaling while preserving long-range contextual fidelity. Building on
this paradigm, this paper investigates substituting standard attention with a
Dynamic Wave recurrent layer that leverages exponential damping and phase
modulation to maintain a stable, long-term memory state.

Existing literature extensively covers linear attention variants
\cite{wang2020linformer,choromanski2021} and pure state-space layer scaling
\cite{gu2023mamba,gu2022s4}. However, the specific behavior of a dynamic wave
mechanism utilizing phase modulation in hybrid configurations remains
insufficiently studied. More importantly, efficiency claims in this literature
are frequently reported without stating which attention backend they are
measured against, even though---as this paper shows---that choice can change
the sign of the result.

This paper bridges these gaps by evaluating a Dynamic Wave layer and its
integration into hybrid language models. Three distinct 10.94M-parameter model
configurations---Pure Transformer, Pure Wave, and Hybrid Wave-Attention---are
formalized and trained on the TinyStories corpus \cite{eldan2023tinystories}.
They are evaluated under a measurement protocol that isolates each
configuration on a clean allocator, separates the cost of the sequence mixer
from that of the shared language-model head, and reports every memory figure
against both a fused and an unfused attention backend.

The main contributions of this paper are as follows:

\begin{itemize}
\item The mathematical apparatus of the Dynamic Wave recurrent layer, based on
input-dependent exponential damping and cosine phase modulation, is formalized
for sequence processing, and its relationship to complex-valued diagonal state
spaces is made explicit.
\item The recurrence is shown to be an associative scan, so that it admits the
same $O(\log T)$-depth parallelization as the state-space layers it is
compared against.
\item Three comparable 10.94M-parameter architectures (Pure Transformer, Pure
Wave, and Hybrid Wave-Attention) are proposed, implemented, and compared at a
matched parameter budget.
\item A measurement protocol is introduced that separates mixer memory from
output-layer memory and controls for the attention backend, the compilation
setting, and allocator state; it is shown that omitting these controls
produces qualitatively wrong conclusions.
\item An implementation-level ablation quantifies how much of the recurrent
layer's latency penalty is architectural and how much is an artifact of an
unoptimized loop.
\item A closed-form model of peak activation memory is derived and validated
against measurement; it reproduces the observed figures and predicts the
context length beyond which attention, rather than the vocabulary projection,
dominates memory.
\item A hardware-aware strategy selection model is developed to guide
architectural choices under explicit operational constraints.
\end{itemize}

% =====================================================================
\section{Related Work}\label{sec:related}

\subsection{Quadratic Attention and Its Hardware Cost}

Multi-Head Attention computes pairwise interactions between every pair of
tokens in a sequence, giving the Transformer its characteristic
$O(T^{2}\!\cdot\!d)$ time and $O(T\!\cdot\!d)$ key-value (KV) cache memory
cost, where $T$ is the sequence length \cite{vaswani2017}. FlashAttention and
its successors reformulate the same computation to be IO-aware, avoiding
materialization of the full $T\times T$ attention matrix in slow GPU memory
and thereby reducing wall-clock cost and peak memory in practice
\cite{dao2022,dao2024flash2,shah2024flash3}. These kernels reduce the constant factor of
attention rather than its asymptotic scaling, and the KV cache still grows
linearly with context length during autoregressive generation. As
Section~\ref{subsec:extreme} shows, however, that constant factor is large
enough to dominate every memory comparison reported in this paper. Tay et al.
provide a broad survey of this cost and the resulting search for sub-quadratic
alternatives \cite{tay2022}.

\subsection{Sparse and Low-Rank Attention Approximations}

A first family of alternatives keeps the attention mechanism but restricts or
approximates which token pairs interact. Longformer combines a local sliding
window with a small number of global tokens \cite{beltagy2020}. Linformer
projects the key and value sequences onto a lower-rank subspace, reducing
complexity to $O(T)$ at the cost of a fixed projection length
\cite{wang2020linformer}. Reformer replaces full attention with
locality-sensitive hashing to approximate nearest-neighbor attention
\cite{kitaev2020}. Performers approximate the softmax kernel with random
feature maps to obtain a linear-time estimator of standard attention
\cite{choromanski2021}. Ring Attention distributes exact attention across
devices using blockwise computation to scale context length rather than
reducing its asymptotic cost \cite{liu2024ring}. These methods remain
attention-shaped: they still model pairwise or approximately pairwise token
interaction, and most require either a specialized kernel or an accuracy
trade-off to realize their theoretical savings.

\subsection{State-Space and Purely Recurrent Alternatives}

A second family removes pairwise interaction entirely and replaces it with a
compressed recurrent state. Mamba introduces an input-dependent
(``selective'') state-space model whose transition parameters are computed
from the input at every step, combined with a hardware-aware parallel scan
\cite{gu2023mamba}. RWKV reformulates attention as a linear, RNN-like
recurrence that can be trained in parallel and run in constant memory during
generation \cite{peng2023rwkv}. RetNet introduces an explicit multi-scale
exponential decay (``retention'') term as a direct algebraic substitute for
softmax attention \cite{sun2023retnet}. Katharopoulos et al. show that linear
attention itself can be reformulated as a recurrent network, unifying the
attention and RNN views \cite{katharopoulos2020}. S4 models long sequences
with a structured, continuous-time state space discretized into a linear
recurrence \cite{gu2022s4}; its diagonal successors S4D \cite{gu2022s4d} and the
Linear Recurrent Unit \cite{orvieto2023lru} keep a per-channel complex
eigenvalue $\lambda = \exp(-\nu + i\theta)$, whose real part damps the state and whose
imaginary part rotates it, so that a damped oscillation is already the
standard behavior of a diagonal state-space channel. Hyena replaces attention with long implicit
convolutions \cite{poli2023hyena}, and H3 combines two state-space layers with
a multiplicative gate to approximate the recall behavior of attention
\cite{fu2023h3}. More recent work shows that selective state spaces and linear
attention are two views of one structured matrix family \cite{dao2024mamba2},
and combines gating with the delta update rule to improve recall at a fixed
state size \cite{yang2025gdn}. Across this family, the recurring engineering requirement is
a custom parallel-scan or convolutional kernel: without one, a sequential
recurrence executed step-by-step in a standard deep-learning framework does
not parallelize over the time dimension the way a matrix multiplication does.
Section~\ref{subsec:short} quantifies how much of that penalty is intrinsic to
the sequential dependency and how much is merely unoptimized work placed
inside the loop.

\subsection{Hybrid Sequence-Mixing Architectures}

A third, more recent direction interleaves recurrent or convolutional layers
with a smaller number of attention layers rather than replacing attention
outright. Mega augments attention with a multi-scale exponential moving
average to inject a recency bias \cite{ma2023mega}. Griffin interleaves gated
linear recurrences with local attention blocks \cite{de2024griffin}. Jamba
scales a transformer-SSM hybrid to production size, alternating Mamba and
attention blocks in a fixed ratio \cite{lieber2024jamba}. A controlled
comparison at a scale of eight billion parameters finds a hybrid with a small
share of attention layers ahead of both pure forms \cite{waleffe2024}, and a
stack in which one softmax attention layer follows every seven linear-attention
layers has been trained at the scale of hundreds of billions of parameters
\cite{minimax2025}. Attention itself was originally introduced as a complement to recurrent encoder--decoder models
rather than as a replacement for them \cite{bahdanau2015}. These results
establish hybridization as a productive design space, typically placing
recurrent or local layers to perform cheap, high-frequency sequence
compression and reserving attention layers for global associative access over
the compressed representation. The Hybrid Wave-Attention configuration
evaluated here follows the same placement principle
(Section~\ref{sec:architectures}).

\subsection{Recall Limitations and the Research Gap}

Arora et al. show empirically that linear-time sequence models, despite
matching attention on perplexity, systematically underperform standard
attention on associative recall sub-tasks, and trace this gap to the
fixed-size compressed state that such models must maintain
\cite{arora2023zoology}. This motivates treating a pure recurrent replacement
of attention as one point in a design space rather than as a drop-in
substitute. The Dynamic Wave layer studied here sits in this design space as a
real-valued formulation of damped oscillation: where a diagonal state space
obtains it from a fixed complex eigenvalue \cite{gu2022s4d,orvieto2023lru},
the Wave layer obtains it from a real damping term and an explicit cosine gate
whose rate, frequency and phase are all functions of the current token, in the
selective spirit of \cite{gu2023mamba}.

What the surveyed literature does not provide is measured---rather than
asymptotic---memory behavior for such layers under a protocol that separates
the cost of the sequence mixer from that of the shared output layer, names the
attention implementation being compared against, and isolates each
configuration from the others on the device. That gap is the subject of this
paper: it formalizes the Dynamic Wave layer
(Section~\ref{sec:wave}) and evaluates it, and a hybrid built from it, under
such a protocol (Sections~\ref{sec:setup}--\ref{sec:results}).

% =====================================================================
\section{Proposed Dynamic Wave Layer}\label{sec:wave}

\subsection{Design Rationale}\label{subsec:rationale}

Retention-based and real-valued state-space layers such as RetNet and Mamba
replace attention with a leaky integrator whose memory decays monotonically
over time \cite{gu2023mamba,sun2023retnet}. Such a kernel can express recency
bias but not constructive or destructive interference between tokens that are
not adjacent, since a purely decaying, non-oscillatory kernel weights every
past token with the same sign. Diagonal state spaces recover oscillation by
moving to complex eigenvalues \cite{gu2022s4d,orvieto2023lru}, at the cost of
complex-valued state and arithmetic. The Dynamic Wave layer keeps the state
real and obtains the same qualitative behavior directly: it retains the
exponential damping envelope, so that recent tokens are still weighted more
heavily than distant ones, but modulates the contribution of each incoming
token with a learned, input-dependent cosine term. Each channel of the hidden state therefore behaves like a driven, damped
harmonic oscillator rather than a monotonically decaying accumulator: the sign
and magnitude with which a token enters the running memory depends on a
learned phase, so that subsequent tokens can reinforce or cancel earlier
contributions. Both the damping rate and the oscillation frequency and phase
are computed from the input at every time step, making the recurrence
selective in the same sense as Mamba's input-dependent transition parameters
\cite{gu2023mamba}, while remaining a much simpler, single-layer recurrence
than a full structured state-space model.

\subsection{Mathematical Formalization}\label{subsec:math}

Let $X = (x_1, \dots, x_T)$ with $x_t \in \mathbb{R}^{d}$ denote the
layer-normalized input sequence of hidden dimension $d$. Three independent
linear projections produce time-varying, input-dependent gate parameters for
every channel:
\begin{align}
\alpha_t &= \mathrm{softplus}(x_t W_\alpha + b_\alpha), \nonumber\\
\omega_t &= x_t W_\omega + b_\omega, \label{eq:gates}\\
\varphi_t &= x_t W_\varphi + b_\varphi, \nonumber
\end{align}
where $\alpha_t > 0$ is the channel-wise forgetting rate, $\omega_t$ is an
instantaneous angular frequency and $\varphi_t$ a phase offset. The hidden
state $h_t \in \mathbb{R}^{d}$ is then updated by a single-step, per-channel
recurrence with $h_0 = 0$:
\begin{equation}
h_t = \exp(-\alpha_t) \odot h_{t-1} + \cos(\omega_t + \varphi_t) \odot x_t,
\label{eq:recur}
\end{equation}
where $\odot$ denotes the Hadamard (element-wise) product. The first term
exponentially damps the previous state channel-by-channel; the second injects
the current input, scaled by a cosine carrier whose frequency and phase are
themselves functions of that same input. Unrolling the recurrence shows that
the effective contribution of an earlier token $x_k$ to the state at a later
step $t$ is scaled by the product of all intermediate damping factors and by
$\cos(\omega_k + \varphi_k)$ evaluated at the time $x_k$ entered the
recurrence---an exponentially decaying, phase-modulated memory kernel rather
than a purely decaying one. Finally, the sequence of states is projected back
to the model dimension by a linear output map:
\begin{equation}
y_t = W_o h_t + b_o .
\label{eq:out}
\end{equation}

\subsection{Parallel Form as an Associative Scan}\label{subsec:scan}

Because $\alpha_t$, $\omega_t$ and $\varphi_t$ depend only on $x_t$ and not
on $h_{t-1}$, the coefficients of \eqref{eq:recur} are known before the
recurrence begins. Writing $a_t = \exp(-\alpha_t)$ and
$b_t = \cos(\omega_t + \varphi_t) \odot x_t$, both obtainable for every
timestep in one batched operation, \eqref{eq:recur} is the first-order linear
recurrence
\begin{equation}
h_t = a_t \odot h_{t-1} + b_t .
\label{eq:linrec}
\end{equation}
Define on pairs of vectors the operator
\begin{equation}
(a_i, b_i) \bullet (a_j, b_j) = (a_i \odot a_j,\; a_j \odot b_i + b_j) .
\label{eq:scanop}
\end{equation}
This operator is associative: both bracketings of a triple reduce to
$(a_1 \odot a_2 \odot a_3,\; a_3 \odot (a_2 \odot b_1 + b_2) + b_3)$,
and $(\mathbf{1}, \mathbf{0})$ is its identity. With $h_0 = 0$, unrolling
\eqref{eq:linrec} gives
\begin{equation}
h_t = \pi_2\!\left[(a_1, b_1) \bullet (a_2, b_2) \bullet \cdots \bullet
(a_t, b_t)\right],
\label{eq:scan}
\end{equation}
where $\pi_2$ selects the second component. The whole state sequence is
therefore the prefix scan of $\{(a_t, b_t)\}_{t=1}^{T}$ under $\bullet$, so a
work-efficient parallel scan \cite{blelloch1990} evaluates all $T$ states in
$O(\log T)$ sequential depth and $O(T\!\cdot\!d)$ work---the same
construction used to parallelize linear recurrent networks
\cite{martin2018} and the selective state-space layers of
Section~\ref{sec:related} \cite{gu2023mamba,gu2022s4}.

Two consequences matter for what follows. First, the sequential dependency in
\eqref{eq:recur} is not an obstacle in principle: the layer admits exactly the
same parallelization as the state-space layers it is compared against.
Second, and independently of any scan, the fact that $a_t$ and $b_t$ are
precomputable means they can simply be hoisted out of a sequential loop,
leaving one multiply--add per step inside it. This restructuring changes
neither the mathematics nor the parameter count, and
Section~\ref{subsec:short} measures what it is worth: most of the layer's
latency penalty, without writing a kernel at all.

\subsection{Integration into a Transformer Block}\label{subsec:block}

The Dynamic Wave layer is used as a drop-in replacement for the mixer
sub-layer of a standard pre-norm Transformer block, so that Wave and Attention
blocks can be freely interleaved in the same backbone. For layer $l$ with
mixer function $\mathrm{Mixer}_l$ (either Multi-Head Attention or the Dynamic
Wave recurrence):
\begin{align}
u^{(l)} &= x^{(l)} + \mathrm{Mixer}_l\!\left(\mathrm{LN}_1(x^{(l)})\right),\\
x^{(l+1)} &= u^{(l)} + \mathrm{FFN}\!\left(\mathrm{LN}_2(u^{(l)})\right),
\end{align}
where $u^{(l)}$ is the intermediate state after the mixer sub-layer,
$\mathrm{FFN}(x) = W_2\,\mathrm{GELU}(W_1 x)$ with the smooth activation of
Hendrycks and Gimpel \cite{hendrycks2016}, and pre-normalization is placed
before rather than after each sub-layer following Xiong et al.
\cite{xiong2020}. Token embeddings are tied with the output (language-model
head) projection. This block-level formalism is shared, unchanged, by all
three configurations; the only difference between them is which mixer is
instantiated at each layer index $l$. Both mixers carry exactly $3d^2+3d$
projection parameters plus a $d^2+d$ output map, so the three configurations
are parameter-matched by construction.

\subsection{Complexity Analysis}\label{subsec:complexity}

Table~\ref{tab:complexity} compares the asymptotic time and memory cost of the
two mixers. ``Generation memory'' refers to the memory required to produce one
additional token given the tokens already generated: for attention this is the
size of the KV cache, which grows with context length; for the Dynamic Wave
recurrence it is, in principle, only the fixed-size state $h_t$, independent
of $T$. The training-memory row for attention is stated twice, because it is
not a property of the mathematics but of the kernel: a fused, IO-aware
implementation never materializes the $T \times T$ score matrix and is
therefore $O(T\!\cdot\!d)$, whereas an unfused implementation is $O(T^2)$.
Section~\ref{sec:results} shows that this distinction, rather than the choice
of mixer, determines every memory ranking reported in this paper.

\begin{table*}[!t]
\caption{Asymptotic Complexity of the Two Mixers}
\label{tab:complexity}
\centering
\begin{tabular}{|l|c|c|c|}
\hline
\textbf{Mixer} & \textbf{Training time} & \textbf{Training memory} &
\textbf{Generation memory (per step, ideal)}\\
\hline
Multi-Head Attention, unfused \cite{vaswani2017} & $O(T^{2}\!\cdot\!d)$ &
$O(T^{2} + T\!\cdot\!d)$ & $O(T\!\cdot\!d)$ (KV cache)\\
\hline
Multi-Head Attention, fused \cite{dao2022} & $O(T^{2}\!\cdot\!d)$ &
$O(T\!\cdot\!d)$ & $O(T\!\cdot\!d)$ (KV cache)\\
\hline
Dynamic Wave (proposed) & $O(T\!\cdot\!d)$ & $O(T\!\cdot\!d)$ &
$O(d)$ (fixed state), not realized---see Section~\ref{subsec:gen}\\
\hline
\end{tabular}
\end{table*}

% =====================================================================
\section{Model Architectures}\label{sec:architectures}

\subsection{Shared Configuration}\label{subsec:shared}

Three model configurations are trained and compared, all sharing an identical
backbone, hyper-parameters, and parameter budget, so that any difference in
behavior can be attributed to the choice and placement of the sequence-mixing
layer rather than to model capacity. All three use hidden dimension $d = 384$,
feed-forward dimension 1536, 6 attention heads (where attention is used), 4
transformer blocks, and a 10\,000-token BPE vocabulary, giving 10.94M
trainable parameters for every configuration
(Table~\ref{tab:configs}). Embedding and language-model head weights are tied.

\begin{table*}[!t]
\caption{Shared Hyper-Parameters and Model Configurations}
\label{tab:configs}
\centering
\begin{tabular}{|l|c|c|c|c|c|}
\hline
\textbf{Configuration} & \textbf{Layer 1} & \textbf{Layer 2} &
\textbf{Layer 3} & \textbf{Layer 4} & \textbf{Parameters}\\
\hline
Pure Transformer & Attention & Attention & Attention & Attention & 10.94M\\
\hline
Pure Wave & Wave & Wave & Wave & Wave & 10.94M\\
\hline
Hybrid Wave-Attention & Wave & Wave & Attention & Attention & 10.94M\\
\hline
\end{tabular}
\end{table*}

\subsection{Pure Transformer (Baseline)}\label{subsec:baseline}

The baseline stacks four identical blocks, each using causal Multi-Head
Attention as the mixer. This configuration is used both as the accuracy
reference and as the only configuration compiled with \texttt{torch.compile}
during training (Section~\ref{subsec:training}).

\subsection{Pure Wave}\label{subsec:purewave}

The Pure Wave configuration replaces every mixer with the Dynamic Wave
recurrence of Section~\ref{sec:wave}, keeping every other component---feed-forward
blocks, normalization, embedding tying, optimizer, and schedule---identical to
the baseline. This isolates the effect of removing attention entirely.

\subsection{Hybrid Wave-Attention}\label{subsec:hybrid}

\begin{figure*}[!t]
\centering
\includegraphics[width=0.92\textwidth]{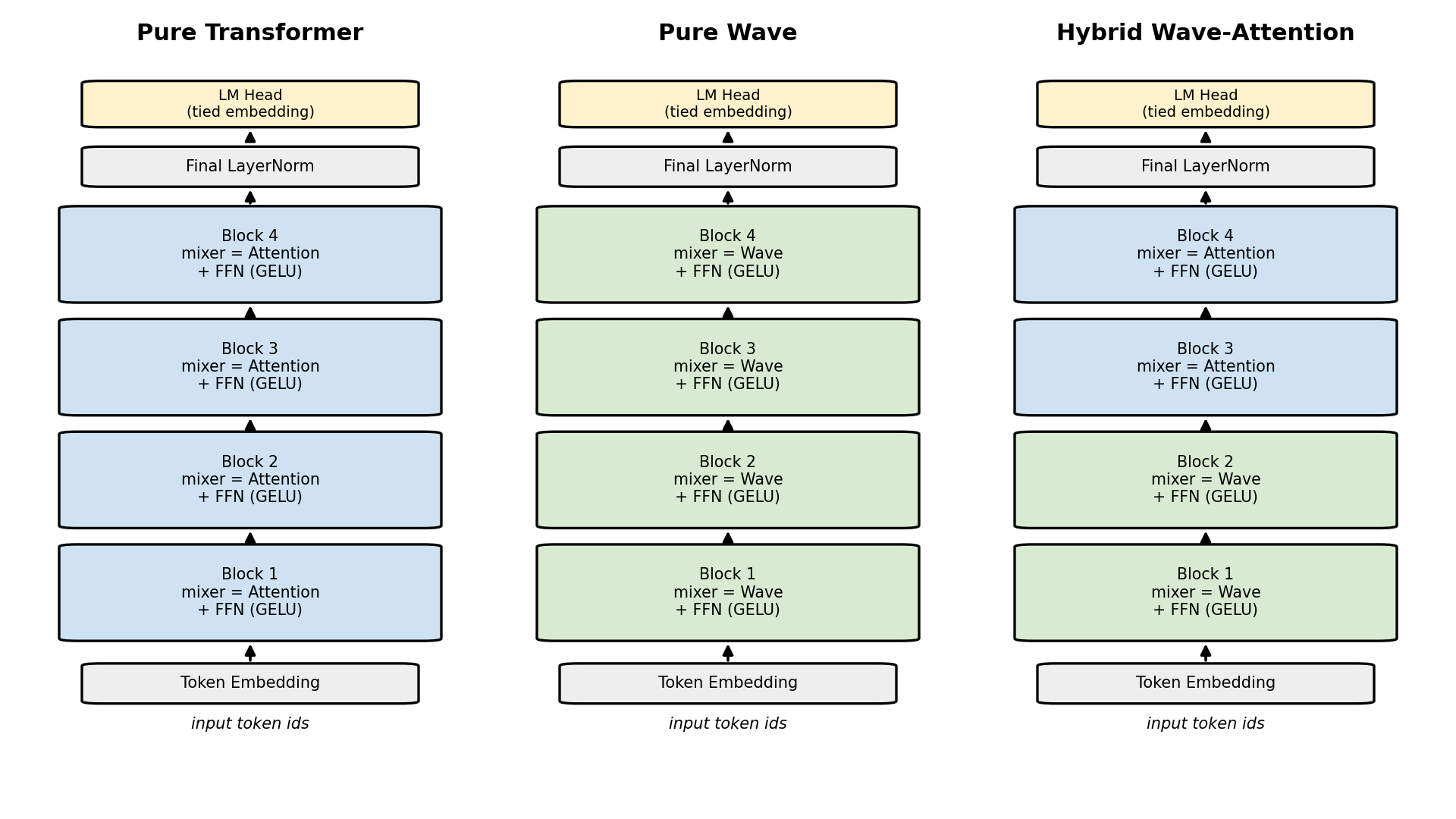}
\caption{Layer configurations of the three evaluated 10.94M-parameter models.}
\label{fig:stacks}
\end{figure*}

\begin{figure*}[!t]
\centering
\includegraphics[width=0.78\textwidth]{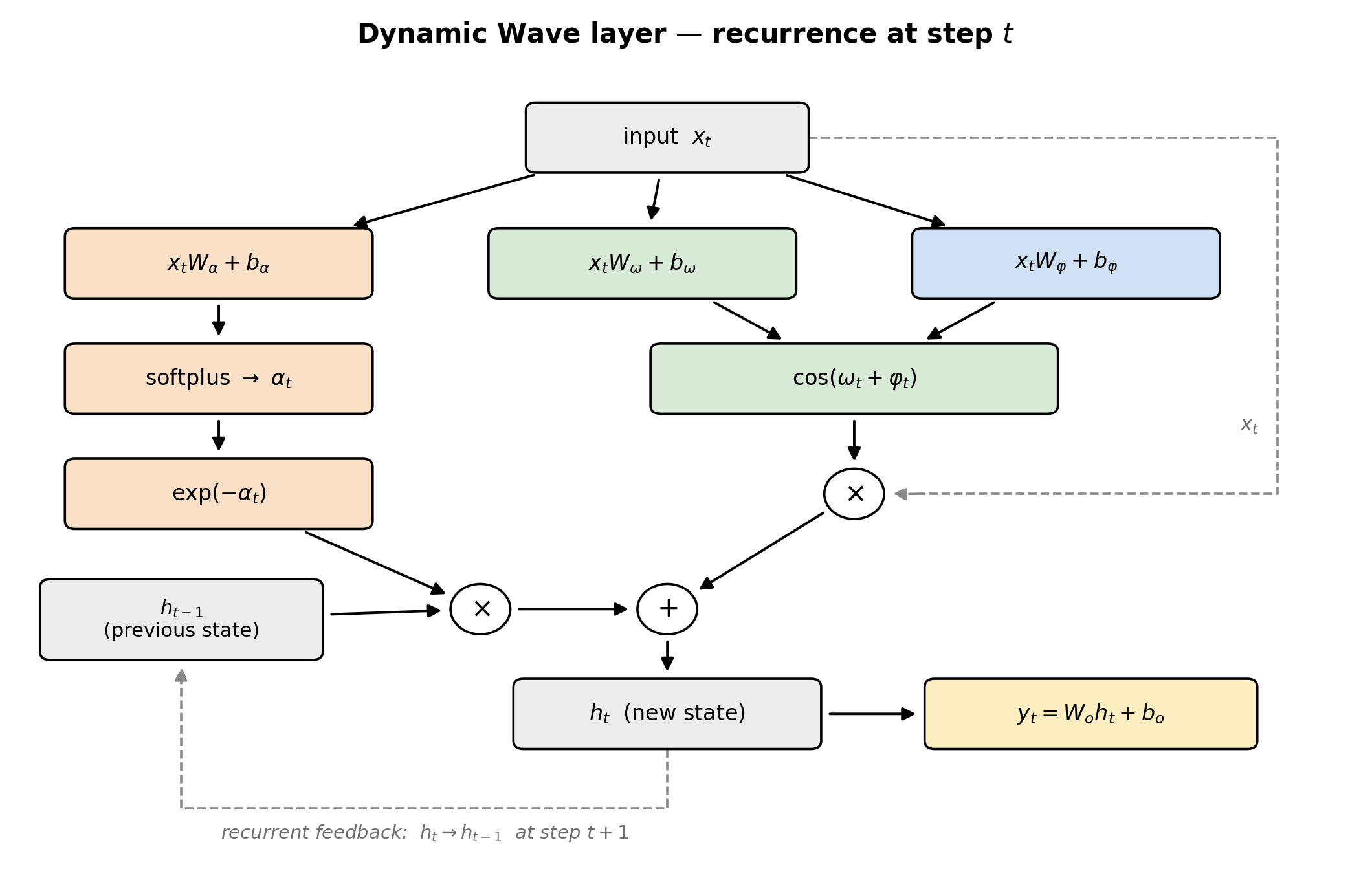}
\caption{Computation graph of the Dynamic Wave layer at step $t$.}
\label{fig:wavegraph}
\end{figure*}

The Hybrid configuration places Wave layers first (blocks 1--2) and Attention
layers last (blocks 3--4). This ordering follows the placement principle
observed in Griffin and Jamba \cite{de2024griffin,lieber2024jamba}: the lower,
Wave-based blocks perform cheap, sequential compression of local context into
a running state, while the upper, Attention-based blocks perform global
associative refinement on top of an already-compressed representation, rather
than needing to model long-range dependence from raw token embeddings
directly. Fig.~\ref{fig:stacks} illustrates the layer stack of all three
configurations, and Fig.~\ref{fig:wavegraph} the internal single-step
computation of the Dynamic Wave layer described by \eqref{eq:gates}--\eqref{eq:out}.

% =====================================================================
\section{Evaluation Metrics}\label{sec:metrics}

\subsection{Perplexity}\label{subsec:pplmetric}

Language-modeling quality is measured by perplexity on held-out TinyStories
chunks. For a batch $b$ with per-token cross-entropy loss $\mathrm{CE}_b$
(averaged over tokens in the batch), the mean loss over $B$ evaluated batches
and the corresponding perplexity are
\begin{equation}
\mathcal{L} = \frac{1}{B}\sum_{b=1}^{B}\mathrm{CE}_b ,
\end{equation}
\begin{equation}
\mathrm{PPL} = \exp(\mathcal{L}).
\end{equation}
Two evaluation regimes are reported: a quick estimate over the first 50
evaluation batches, and a full evaluation over all 63 batches of the held-out
split (2000 chunks not seen during training).

\subsection{Token Generation Throughput}\label{subsec:throughputmetric}

Throughput is measured as generated tokens per second under greedy (arg-max)
decoding, in two regimes that isolate the effect of context growth. In the
bounded-window regime, the model generates 100 tokens while the context fed to
the model at each step is truncated to the last 512 tokens, matching the
sequence length used during training. In the unbounded, growing-context
regime, the model generates 1500 tokens while the full, ever-growing context
is re-fed at every step without truncation.

\subsection{Peak VRAM Profiling}\label{subsec:vrammetric}

Peak GPU memory is measured with \texttt{torch.cuda.\allowbreak max\_\allowbreak memory\_\allowbreak allocated}.
Four controls are applied, each of which was found to be necessary to obtain
architecture-attributable numbers:

\begin{enumerate}
\item \textit{Isolation.} Each configuration is constructed, measured and
destroyed on its own, with an explicit garbage collection and cache release
before and after, so that no other model's weights or allocator state
contribute to its peak. Measuring the three configurations while all of them
remain resident inflates every figure and, worse, makes the last-measured
configuration inherit the most fragmented allocator.
\item \textit{Activation memory.} The reported quantity is
$\text{peak} - \text{steady state}$, where the steady state is measured after
warm-up with only parameters resident. This removes the constant weight
footprint from the comparison.
\item \textit{Output-layer separation.} Every sweep is run twice, once with
the tied language-model head and once stopping before it. The
$[B, T, V]$ logit tensor is the single largest allocation in the full model
and is identical across the three configurations by construction, so it masks
the mixers entirely unless removed (Section~\ref{subsec:short}).
\item \textit{Named attention backend.} Every attention measurement is
reported for both an unfused implementation, which materializes the
$[B, H, T, T]$ score matrix, and a fused scaled-dot-product implementation,
which does not.
\end{enumerate}

Throughout, \textit{reference implementation} denotes the code used to train the three models and to obtain the throughput and wall-clock results: causal attention through \texttt{nn.\allowbreak MultiheadAttention} with an explicit floating-point mask, and the Dynamic Wave recurrence with $\exp(-\alpha_t)$ and $\cos(\omega_t+\varphi_t)$ evaluated inside the per-step loop. \textit{Optimized implementation} denotes the same two mixers with attention routed through a fused scaled-dot-product kernel and the Wave recurrence restructured as described in Section~\ref{subsec:math}. The two differ only in how the same mathematics is executed; parameter counts and outputs are identical.

A short-context sweep measures peak VRAM and average forward-pass latency at
sequence lengths of 128, 256 and 512 tokens with batch size 16, after warm-up,
averaged over 30 timed iterations. An extreme-context sweep measures peak VRAM
at 1024, 2048 and 4096 tokens with batch size 8. A separate sweep measures
memory in training mode, with a backward pass, at 512 tokens and batch size 8.

% =====================================================================
\section{Experimental Setup}\label{sec:setup}

\subsection{Dataset and Tokenization}\label{subsec:data}

All models are trained on a 500\,000-story subset of the TinyStories corpus
\cite{eldan2023tinystories}, a dataset of short, syntactically simple
children's stories designed to let small models learn coherent English. A
byte-pair-encoding tokenizer with a 10\,000-token vocabulary is trained from
scratch on the same subset using whitespace pre-tokenization. Tokenized
stories are concatenated and split into fixed-length chunks of 512 tokens,
yielding 207\,113 training chunks; the last 2000 chunks are held out for
evaluation and are not seen during training. Training batches of 32 chunks are
sampled with shuffling and dropped-last batching, giving 6472 optimization
steps per epoch.

\subsection{Training Configuration}\label{subsec:training}

All three configurations are trained for one epoch (6472 steps) with the AdamW
optimizer (weight decay 0.01) \cite{loshchilov2019}, using a One-Cycle
learning-rate schedule with cosine annealing, a peak learning rate of
$5\times10^{-4}$, and a 500-step warm-up---the super-convergence recipe of
Smith and Topin \cite{smith2019}. Training uses mixed-precision autocast with
gradient scaling \cite{micikevicius2018} and gradient-norm clipping at 1.0.
All models are implemented in PyTorch \cite{paszke2019}; the Pure Transformer
baseline is additionally compiled with \texttt{torch.compile}, which lowers
the model through TorchInductor onto Triton GPU kernels
\cite{ansel2024pytorch2,tillet2019triton}, while the Pure Wave and Hybrid
models are trained in eager mode, because \texttt{torch.compile} does not, at
the time of writing, handle the explicit per-timestep Python loop of the
Dynamic Wave recurrence. This asymmetry affects the training wall-clock comparison of
Section~\ref{subsec:wallclock} and the generation-throughput comparison of
Section~\ref{subsec:gen}, both of which were obtained with the compiled
baseline. The memory and forward-latency benchmarks of
Sections~\ref{subsec:short}--\ref{subsec:trainmem} are run uncompiled for all
three configurations and are therefore free of it.

All experiments run on a single NVIDIA Tesla T4 GPU (16\,GB GDDR6, Turing,
compute capability 7.5) in a Kaggle notebook environment with Python 3.12.12,
PyTorch 2.10.0 and CUDA 12.8. Because the T4 does not support BF16,
mixed-precision training uses FP16 autocast with gradient scaling; all
profiling runs use FP32.

\subsection{Research Questions}\label{subsec:rq}

\begin{itemize}
\item \textbf{RQ1:} At a fixed parameter budget, does replacing Multi-Head
Attention with the Dynamic Wave recurrence change language-modeling quality,
and does a hybrid placement improve on both pure configurations?
\item \textbf{RQ2:} Does the theoretical $O(T)$ memory advantage of a
recurrent mixer materialize in practice, and does the answer depend on the
attention backend it is compared against?
\item \textbf{RQ3:} How much of the recurrent mixer's measured latency penalty
is attributable to the sequential dependency itself, and how much to
unoptimized work placed inside the loop?
\item \textbf{RQ4:} Does a hybrid stack inherit the memory behavior of its
cheaper or of its more expensive mixer?
\end{itemize}

% =====================================================================
\section{Results}\label{sec:results}

\subsection{Language-Modeling Quality}\label{subsec:quality}

Table~\ref{tab:quick} reports the quick, 50-batch perplexity estimate together
with the short-context latency and memory sweep;
Table~\ref{tab:fulleval} reports the full, 63-batch evaluation over the entire
held-out split. Both evaluations rank the three configurations identically:
Hybrid Wave-Attention achieves the lowest (best) perplexity, Pure Transformer
is second, and Pure Wave is third. On the full evaluation, Hybrid reaches a
perplexity of 5.95 (average cross-entropy loss 1.7831) versus 6.93 (loss
1.9357) for the Transformer baseline and 7.26 (loss 1.9818) for Pure Wave---a
14.1\% perplexity reduction relative to the baseline and 18.0\% relative to
Pure Wave. The quick 50-batch estimate (6.75 / 7.05 / 5.79) is consistent with
this ranking and differs from the full evaluation by less than 0.3 perplexity
points for every configuration.

\begin{table*}[!t]
\caption{Quick Benchmark: Quality, Forward Latency, and Short-Context
Activation Memory (batch 16)}
\label{tab:quick}
\centering
\begin{tabular}{|l|c|c|c|c|c|c|c|}
\hline
\textbf{Model} & \textbf{PPL} &
\multicolumn{3}{c|}{\textbf{Latency, reference impl. (ms)}} &
\textbf{Latency @512} &
\multicolumn{2}{c|}{\textbf{Activation memory @512 (MB)}}\\
\cline{3-5}\cline{7-8}
& \textbf{(quick)} & \textbf{128} & \textbf{256} & \textbf{512} &
\textbf{optimized (ms)} & \textbf{full model} & \textbf{mixer only}\\
\hline
Pure Transformer & 6.75 & 15.8 & 32.8 & 72.6 & 63.3 & 324.5 & 265.0\\
\hline
Pure Wave & 7.05 & 46.1 & 89.2 & 180.2 & 71.2 & 324.5 & 120.0\\
\hline
Hybrid Wave-Attention & 5.79 & 24.1 & 50.0 & 114.2 & 67.3 & 324.5 & 265.0\\
\hline
\end{tabular}
\end{table*}

\begin{table}[!t]
\caption{Full-Dataset Evaluation (63 Held-Out Batches)}
\label{tab:fulleval}
\centering
\begin{tabular}{|l|c|c|}
\hline
\textbf{Model} & \textbf{Avg. Cross-Entropy Loss} & \textbf{Perplexity}\\
\hline
Pure Transformer & 1.9357 & 6.93\\
\hline
Pure Wave & 1.9818 & 7.26\\
\hline
Hybrid Wave-Attention & 1.7831 & 5.95\\
\hline
\end{tabular}
\end{table}

\subsection{Qualitative Generation Behavior}\label{subsec:qualitative}

For the prompt ``Once upon a time, a little boy named Tom went to the forest
and found a'', all three models produce fluent, on-topic continuations under
greedy decoding. The Transformer continues with ``a big tree'', which Tom
climbs; the Hybrid model produces a similar but independently-phrased
continuation about climbing a tree; Pure Wave instead introduces a different
object (``a big, shiny rock'') and a different narrative arc, demonstrating
that Pure Wave is not simply imitating the Transformer's output distribution.
On a second, longer prompt designed to probe entity tracking over a longer
span, the Transformer and Hybrid models both continue coherently and keep
referring to the already-introduced characters, while Pure Wave's continuation
introduces an ungrounded new character not present in the prompt---a mild form
of entity drift consistent with the weaker associative-recall behavior
reported for linear-time sequence models in general \cite{arora2023zoology}. A
temperature- and noise-robustness stress test on the Hybrid model
(temperature 0.2, 1.5, and 10\% random-token corruption of the prompt) shows
graceful degradation in every case.

\subsection{Interpretability: Attention Focus and Wave Memory Dynamics}
\label{subsec:interp}

\begin{figure}[!t]
\centering
\includegraphics[width=0.92\columnwidth]{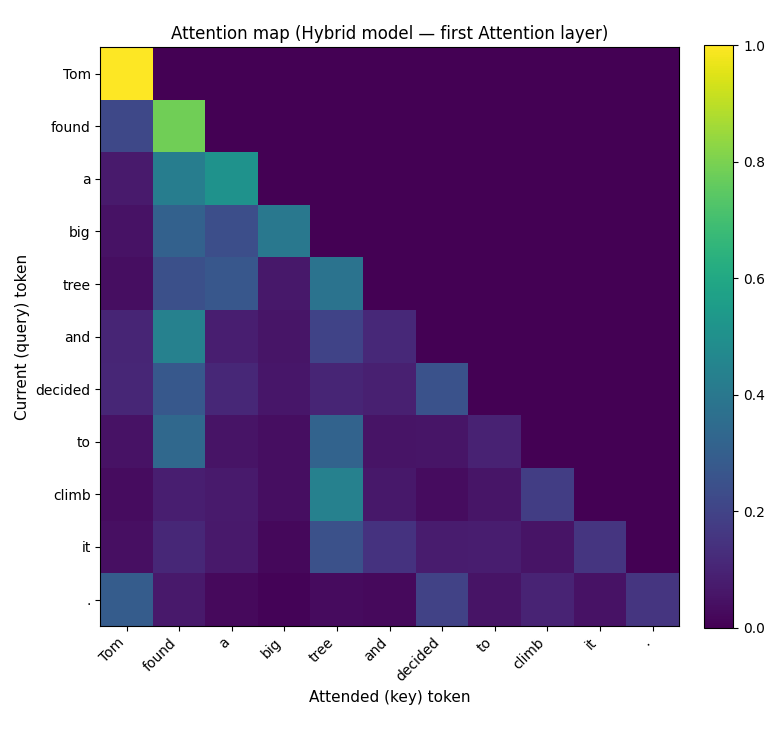}
\caption{Attention map of the first Attention layer in the Hybrid model.}
\label{fig:attnmap}
\end{figure}

\begin{figure}[!t]
\centering
\includegraphics[width=\columnwidth]{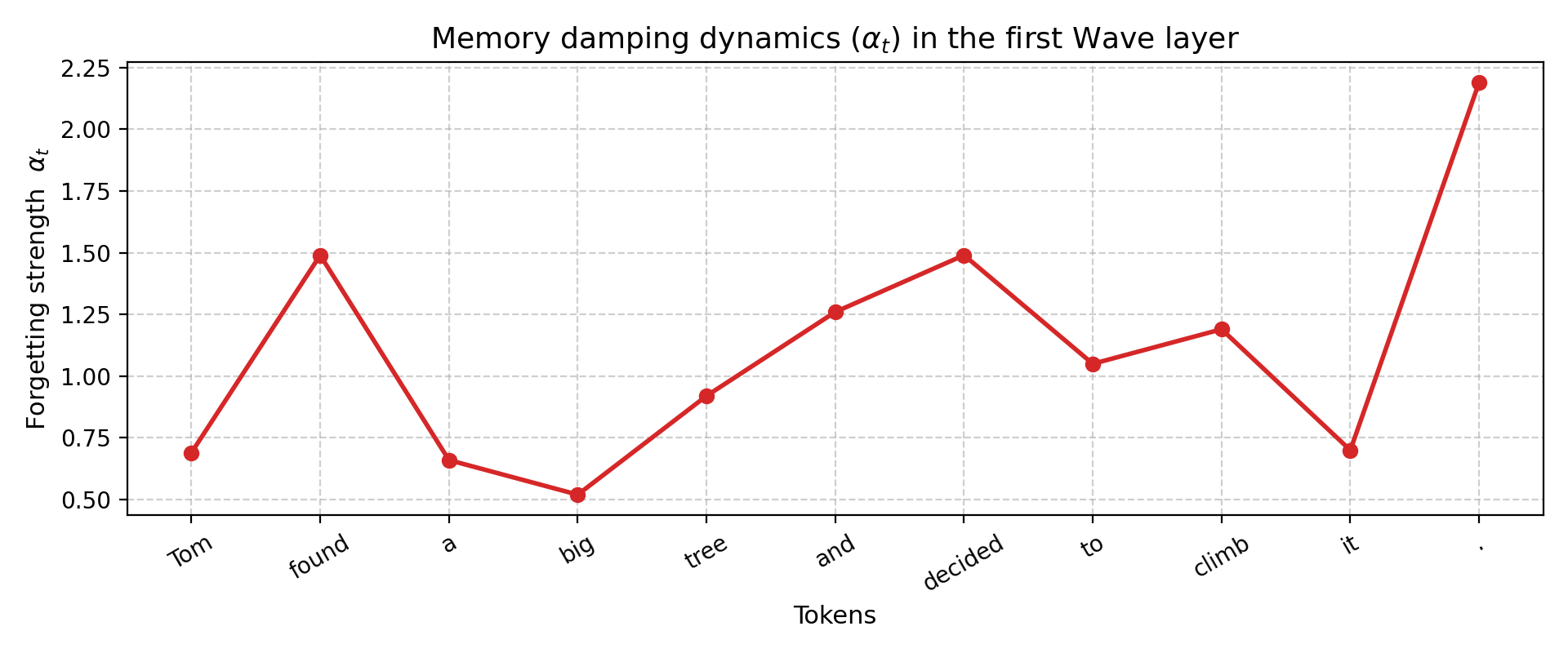}
\caption{Dynamics of the memory damping coefficient $\alpha_t$ in the first
Wave layer of the Hybrid model.}
\label{fig:alpha}
\end{figure}

To inspect how the two mixer types behave inside the trained Hybrid model,
forward hooks were attached to its first Attention layer (block 3) and its
first Wave layer (block 1) while processing the sentence ``Tom found a big
tree and decided to climb it.'' Fig.~\ref{fig:attnmap} shows the resulting
attention-weight matrix, and Fig.~\ref{fig:alpha} the mean forgetting strength
$\alpha_t$ at every token position.

In the first half of the sentence most of the attention mass in
Fig.~\ref{fig:attnmap} stays on the current token and on the verb ``found''.
Later tokens attend selectively to earlier content words rather than to their
immediate neighbors: ``climb'' and ``it'' place their largest weight on
``tree'' (0.44 and 0.25), and the sentence-final period places it on the
subject ``Tom'' (0.29). The $\alpha_t$ dynamics in Fig.~\ref{fig:alpha} show
that the forgetting gate is not a fixed decay rate: the channel-averaged rate
varies from token to token between 0.53 and 1.49 and reaches its maximum of
2.19 at the sentence-final period, where the state is flushed most strongly---a
content-dependent, selective forgetting behavior reminiscent of Mamba's
input-dependent transition parameters \cite{gu2023mamba}, obtained here from a
much simpler recurrence.

\subsection{Short-Context Speed and Memory Trade-off}\label{subsec:short}

At the short context lengths used during training
(Table~\ref{tab:quick}), activation memory measured on the full model is
\textit{identical} across all three configurations at every length: 81.1, 162.2
and 324.5~MB at 128, 256 and 512 tokens respectively. This is not a property
of the mixers. The peak is set by the $[B, T, V]$ logit tensor produced by the
tied language-model head, which at $T = 512$ with batch 16 in FP32 occupies
312.5~MiB, or 96\% of the measured activation total, and which is identical
across configurations by construction. Repeating the sweep with the head
removed separates the mixers: at 512 tokens the unfused attention stack
allocates 265.0~MB against 120.0~MB for the Wave stack, a factor of 2.2. Any
memory comparison that includes the output layer at these scales measures the
vocabulary projection, not the sequence mixer.

Latency separates the configurations sharply, but the separation is largely an
implementation artifact. With the reference implementations---an unfused
attention path and a Wave recurrence that evaluates $\exp(-\alpha_t)$ and
$\cos(\omega_t + \varphi_t)$ inside the per-step loop---Pure Wave requires
180.2~ms per forward pass at 512 tokens against 72.6~ms for the Transformer, a
2.48$\times$ penalty, with the Hybrid model at 114.2~ms in proportion to its
share of Wave layers. Hoisting the two elementwise terms out of the loop, as
noted in Section~\ref{subsec:math}, leaves the recurrence and the parameter
count unchanged but reduces Pure Wave to 71.2~ms---a 2.53$\times$ speed-up
obtained with no custom kernel, no fused scan, and no change to the
mathematics. Against a fused attention baseline at 63.3~ms, the optimized Wave
layer is then only 1.12$\times$ slower. The sequential dependency itself is
therefore a far smaller obstacle at these sequence lengths than the amount of
work placed inside the sequential region.

\subsection{Extreme-Context Memory Scaling}\label{subsec:extreme}

\begin{figure*}[!t]
\centering
\includegraphics[width=0.92\textwidth]{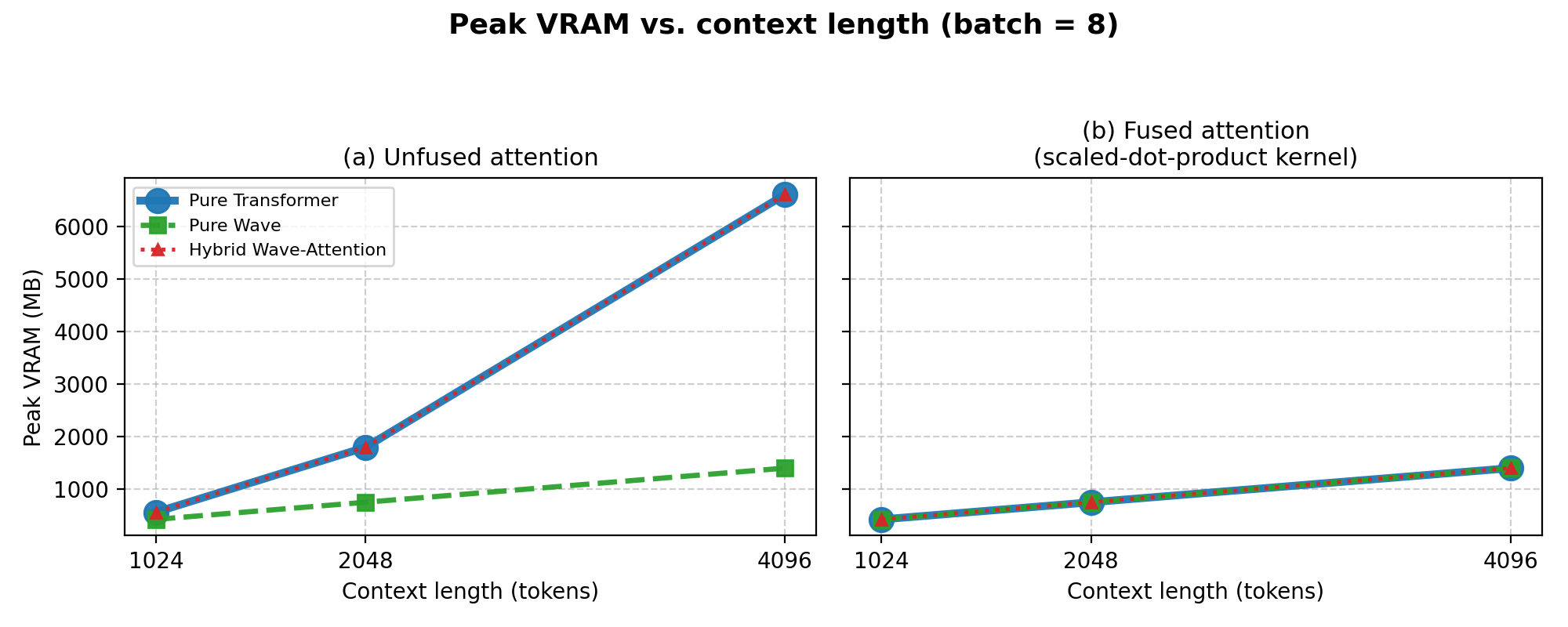}
\caption{Peak VRAM vs. context length at batch 8, under (a) an unfused
attention implementation and (b) a fused scaled-dot-product kernel. The Pure
Wave curve is identical in both panels because it contains no attention. Curves
coincide where the underlying figures do: the Transformer and Hybrid curves
overlie each other in (a), and all three overlie each other in (b), which is
the result reported in Section~\ref{subsec:extreme}.}
\label{fig:vram}
\end{figure*}

Table~\ref{tab:extreme} and Fig.~\ref{fig:vram} report peak VRAM at 1024, 2048
and 4096 tokens with batch size 8, for both attention backends.

Against the unfused baseline the recurrent mixer's advantage is large and
grows with context: at 4096 tokens Pure Wave uses 1404.3~MB against 6601.6~MB
for the Transformer, a 78.7\% reduction, and on a mixer-only basis 480.0~MB
against 6496.0~MB, a 92.6\% reduction. Against a fused scaled-dot-product
kernel the same comparison collapses entirely: 1404.3~MB for Pure Wave against
1403.6~MB for the Transformer, a difference of 0.05\% in the Transformer's
favor, and on a mixer-only basis 480.0~MB for both. The Dynamic Wave layer's
long-context memory advantage is thus not an advantage over attention as such,
but over one particular way of computing attention---one that the field has
already largely abandoned \cite{dao2022,dao2024flash2,shah2024flash3}.

The Hybrid configuration answers RQ4 unambiguously and in the same way under
both backends: its peak matches the Pure Transformer to within 0.1\%
(6602.5~MB vs. 6601.6~MB unfused; 1404.5~MB vs. 1403.6~MB fused). A hybrid
stack inherits the cost of its more expensive mixer, not the advantage of its
cheaper one, because the peak is set by the largest single allocation
anywhere in the forward pass, and replacing half the attention layers with
Wave layers does not reduce that maximum. Halving the number of attention
layers therefore buys nothing in peak memory, although it would reduce
aggregate KV-cache size during incremental decoding.

\begin{table*}[!t]
\caption{Peak VRAM (MB) at Extreme Context Lengths, Batch 8, Language-Model
Head Included}
\label{tab:extreme}
\centering
\begin{tabular}{|l|c|c|c|c|c|c|}
\hline
\textbf{Model} &
\multicolumn{3}{c|}{\textbf{Unfused attention}} &
\multicolumn{3}{c|}{\textbf{Fused attention}}\\
\cline{2-7}
& \textbf{1024} & \textbf{2048} & \textbf{4096}
& \textbf{1024} & \textbf{2048} & \textbf{4096}\\
\hline
Pure Transformer & 565.4 & 1801.5 & 6601.6 & 429.9 & 755.5 & 1403.6\\
\hline
Pure Wave & 430.6 & 756.2 & 1404.3 & 430.6 & 756.2 & 1404.3\\
\hline
Hybrid Wave-Attention & 566.3 & 1802.3 & 6602.5 & 430.8 & 756.3 & 1404.5\\
\hline
\end{tabular}
\end{table*}

\subsection{An Analytical Model of Peak Activation Memory}
\label{subsec:model}

The measurements above are not only reproducible; they are predictable in
closed form, which makes the qualitative conclusions independent of this
particular hardware. In FP32, three tensor families account for essentially
all activation memory in a forward pass:
\begin{align}
A_{\mathrm{logit}} &= 4\,B\,T\,V, \\
A_{\mathrm{attn}}  &= 8\,B\,H\,T^{2}, \\
A_{\mathrm{lin}}   &= 4\,c\,B\,T\,d,
\end{align}
where $B$ is the batch size, $V$ the vocabulary size, $H$ the number of heads,
and $c$ the number of $[B, T, d]$ activations simultaneously live inside a
block. The factor $8$ in $A_{\mathrm{attn}}$ (rather than $4$) reflects that
the unfused path holds the raw score matrix and its softmax output at the same
time; a fused kernel contributes no $T^{2}$ term at all. Fitting $c$ to the
mixer-only sweeps gives $c \approx 10$ for the Wave stack and $c \approx 11$
for the attention stack.

Peak memory is then set by whichever of two moments in the forward pass is
larger: the interior of a block, where the mixer's own activations are live,
or the final projection, where the logit tensor and the last hidden state are
live together:
\begin{equation}
A_{\mathrm{peak}} \approx
\max\!\left(A_{\mathrm{attn}} + A_{\mathrm{lin}},\;
            A_{\mathrm{logit}} + 4\,B\,T\,d\right).
\label{eq:memmodel}
\end{equation}

Table~\ref{tab:memmodel} evaluates \eqref{eq:memmodel} against the measured
configurations. Where the logit branch dominates, the model is exact to the
reported precision; where the quadratic attention branch dominates it is
accurate to a few percent at long context, degrading at short context as the
constant-factor terms it does not model become comparatively larger.

\begin{table*}[!t]
\caption{Predicted \eqref{eq:memmodel} vs. Measured Activation Memory (MB)}
\label{tab:memmodel}
\centering
\begin{tabular}{|l|c|c|c|c|c|}
\hline
\textbf{Configuration} & \textbf{$B$} & \textbf{$T$} & \textbf{Predicted} &
\textbf{Measured} & \textbf{Error}\\
\hline
Wave, with head & 16 & 512 & 324.5 & 324.5 & $0.0\%$\\
\hline
Wave, with head & 8 & 1024 & 324.5 & 324.5 & $0.0\%$\\
\hline
Wave, with head & 8 & 2048 & 649.0 & 650.0 & $-0.2\%$\\
\hline
Wave, with head & 8 & 4096 & 1298.0 & 1298.0 & $0.0\%$\\
\hline
Fused attn., with head & 8 & 4096 & 1298.0 & 1298.0 & $0.0\%$\\
\hline
Fused attn., mixer only & 8 & 4096 & 480.0 & 480.0 & $0.0\%$\\
\hline
Unfused attn., with head & 8 & 2048 & 1776.0 & 1696.0 & $+4.7\%$\\
\hline
Unfused attn., with head & 8 & 4096 & 6624.0 & 6496.0 & $+2.0\%$\\
\hline
\end{tabular}
\end{table*}

The model also predicts, without any further measurement, the context length
at which unfused attention overtakes the output layer as the dominant
allocation. Setting $A_{\mathrm{attn}} > A_{\mathrm{logit}}$ gives
\begin{equation}
T^{*} > \frac{V}{2H},
\label{eq:crossover}
\end{equation}
which for the vocabulary and head count used here is
$T^{*} \approx 10\,000 / 12 \approx 833$ tokens. The measured crossover lies
between 512 tokens, where the Transformer's activation memory equals the Wave
model's at 324.5~MB, and 1024 tokens, where it exceeds it at 460.0~MB against
324.5~MB---as \eqref{eq:crossover} requires.

Equation~\eqref{eq:crossover} also explains the otherwise surprising fused
column of Table~\ref{tab:extreme}. A fused kernel removes $A_{\mathrm{attn}}$
entirely, leaving the logit branch dominant at every length tested, and that
branch is identical for all three configurations by construction. The three
architectures must therefore coincide, and they do, to within 0.1\%. What is
being compared in the fused regime is not the mixers at all but the shared
vocabulary projection---which is the practical reason that a sub-quadratic
mixer buys nothing here, and would begin to matter only for
$T \gg T^{*}$ or for a substantially smaller vocabulary.

\subsection{Training Memory}\label{subsec:trainmem}

Table~\ref{tab:trainmem} reports activation memory for a full training step
(forward and backward) at 512 tokens with batch size 8. The ranking inverts
relative to the unfused inference case. Against the unfused baseline Pure Wave
retains an advantage (621.2~MB against 789.2~MB, 21.3\% less), but against a
fused attention baseline Pure Wave is the \textit{most} expensive of the three
configurations, at 621.2~MB against 573.6~MB, 8.3\% more. The reason is that
the Wave recurrence must retain every intermediate state $h_t$ for the
backward pass---an $O(T\!\cdot\!d)$ autograd tape with $T$ separate nodes---whereas
a fused attention kernel recomputes its intermediates and stores comparatively
little. A recurrent mixer that is cheaper at inference time is therefore not
automatically cheaper to train.

\begin{table}[!t]
\caption{Training Activation Memory, Forward and Backward, 512 Tokens,
Batch 8 (MB)}
\label{tab:trainmem}
\centering
\begin{tabular}{|l|c|c|}
\hline
\textbf{Model} & \textbf{Unfused attention} & \textbf{Fused attention}\\
\hline
Pure Transformer & 789.2 & 573.6\\
\hline
Pure Wave & 621.2 & 621.2\\
\hline
Hybrid Wave-Attention & 705.2 & 597.4\\
\hline
\end{tabular}
\end{table}

\subsection{Inference Throughput: Bounded vs. Unbounded Context}
\label{subsec:gen}

Table~\ref{tab:throughput} reports generation throughput for the reference
implementations in the bounded-window regime (100 tokens generated, context
truncated to the last 512 tokens) and in the unbounded, growing-context regime
(1500 tokens generated, full un-truncated context re-fed at every step). In
the bounded-window regime the Transformer reaches 526.22 tokens/sec against
49.55 for Pure Wave and 80.25 for Hybrid. In the unbounded regime throughput
collapses for the recurrent mixers: the Transformer falls to 102.31
tokens/sec, while Pure Wave falls to 3.96 and Hybrid to 7.75.

This collapse has a different cause from the forward-pass latency discussed in
Section~\ref{subsec:short}, and it is not removed by hoisting work out of the
loop. The implementation re-computes the Dynamic Wave recurrence from the
first token at every decoding step, because no incremental, state-caching
decoding path is implemented: a mechanism that is $O(T)$ per step by
construction is thereby executed as $O(T^2)$ in total. An incremental
interface carrying only the fixed-size state $h_t$ between steps was
scaffolded during this study but is not implemented in the reference model
used here; such a decoding path would remove this cost.

\begin{table*}[!t]
\caption{Generation Throughput: Bounded vs. Unbounded Context
(reference implementations)}
\label{tab:throughput}
\centering
\begin{tabular}{|l|c|c|c|}
\hline
\textbf{Model} & \textbf{Bounded window, 100 tok (tok/s)} &
\textbf{Unbounded context, 1500 tok (tok/s)} & \textbf{Slowdown}\\
\hline
Pure Transformer & 526.22 & 102.31 & 5.1$\times$\\
\hline
Pure Wave & 49.55 & 3.96 & 12.5$\times$\\
\hline
Hybrid Wave-Attention & 80.25 & 7.75 & 10.3$\times$\\
\hline
\end{tabular}
\end{table*}

\subsection{Training Wall-Clock Time}\label{subsec:wallclock}

Table~\ref{tab:wallclock} reports the wall-clock duration of one full training
epoch (6472 steps). The Pure Transformer trains in 18.36 minutes, versus
328.27 minutes for Pure Wave and 174.99 minutes for Hybrid. As discussed in
Section~\ref{subsec:training}, only the Transformer benefits from
\texttt{torch.compile} here, so these figures conflate an architectural cost
with an engineering one and should not be read as an architecture-only
comparison. In the light of Section~\ref{subsec:short}, a substantial part of
the gap is attributable to the unoptimized recurrence rather than to the
sequential dependency: the same 2.53$\times$ forward-pass speed-up would apply
to every training step.

\begin{table}[!t]
\caption{Wall-Clock Training Time, One Epoch (6472 Steps)}
\label{tab:wallclock}
\centering
\begin{tabular}{|l|c|c|}
\hline
\textbf{Model} & \textbf{Training time (min)} & \texttt{torch.compile}\\
\hline
Pure Transformer & 18.36 & Yes\\
\hline
Pure Wave & 328.27 & No\\
\hline
Hybrid Wave-Attention & 174.99 & No\\
\hline
\end{tabular}
\end{table}

\begin{figure*}[!t]
\centering
\includegraphics[width=0.92\textwidth]{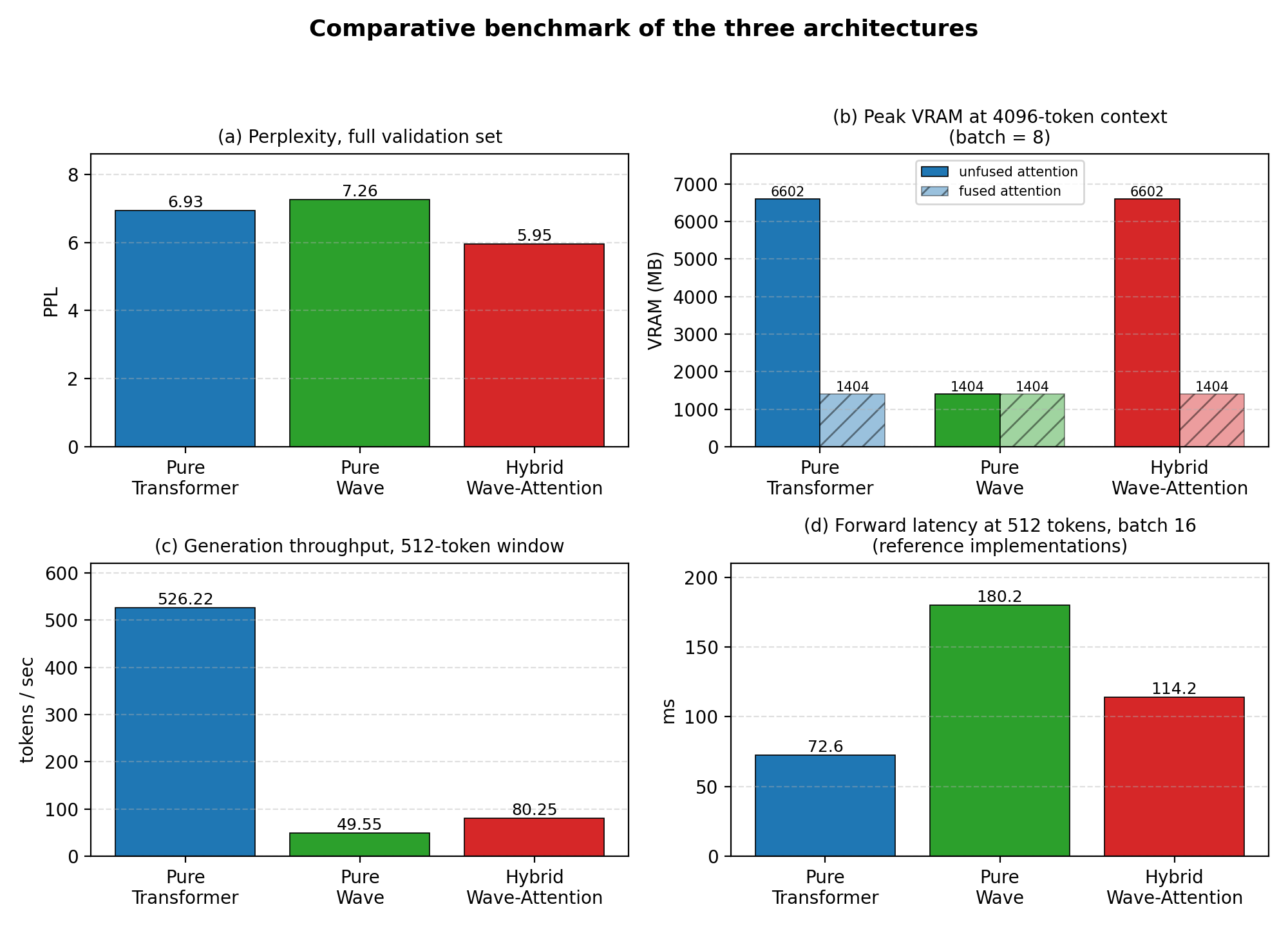}
\caption{Comparative benchmark of the three architectures: perplexity, peak
VRAM at 4096 tokens under both attention backends, generation throughput, and
forward latency.}
\label{fig:summary}
\end{figure*}

% =====================================================================
\section{Discussion}\label{sec:discussion}

\subsection{Hybridization Improves Quality, and Is Neutral for Peak Memory}
\label{subsec:hybridmem}

The Hybrid configuration delivers the best language-modeling quality of the
three (14.1\% lower perplexity than the Transformer baseline,
Section~\ref{subsec:quality}) and the best qualitative entity-tracking
behavior (Section~\ref{subsec:qualitative}), supporting the hypothesis that
placing cheap sequential compression before global attention, as in Griffin
and Jamba \cite{de2024griffin,lieber2024jamba}, is an effective design pattern
at this scale. On peak memory, however, it is indistinguishable from the Pure
Transformer under both attention backends
(Section~\ref{subsec:extreme}). Peak memory is determined by the largest
single allocation in the forward pass, so replacing half of the attention
layers leaves that maximum untouched. A hybrid design should therefore be
justified on the quality axis; on the peak-memory axis it is neither a gain
nor a loss, and its benefit appears only in aggregate quantities such as
KV-cache size during incremental decoding.

The isolation control of the protocol matters in practice. When the three
configurations are measured co-resident on the device in a fixed order, the
Hybrid configuration appears the most memory-hungry, at 8021.9~MB against
4373.9~MB for the Transformer; measured in isolation, the difference disappears
and the results do not depend on the measurement order.

\subsection{Linear Memory Is Real, but Only Against an Unfused Baseline}
\label{subsec:linearmem}

Pure Wave shows a large, sustained memory advantage over unfused attention at
long context---78.7\% at 4096 tokens, and 92.6\% on a mixer-only basis. That
advantage disappears completely against a fused scaled-dot-product kernel,
where the two mixers are equal to within measurement noise, and reverses in
training mode, where the Wave layer's autograd tape makes it 8.3\% more
expensive than fused attention (Section~\ref{subsec:trainmem}).

This is the central methodological finding of this paper. An efficiency claim
for a sub-quadratic sequence mixer is not well-defined unless the attention
implementation it is measured against is named. Since fused, IO-aware attention is the default in current
frameworks, comparisons against an unfused baseline should be regarded as
upper bounds that are unlikely to be realized in deployment.

\subsection{Most of the Latency Penalty Is Not Architectural}
\label{subsec:kernel}

The state-space literature attributes the wall-clock gap between recurrent
mixers and attention to the absence of a parallel scan over the time dimension,
which is why Mamba, S4 and RWKV pair their recurrences with dedicated kernels
\cite{gu2023mamba,peng2023rwkv,gu2022s4}. That account is incomplete for the
layer studied here. Because $\alpha_t$, $\omega_t$ and $\varphi_t$ depend only
on $x_t$, the transcendental terms can be evaluated for all timesteps at once
before the loop begins; doing so speeds the layer up 2.53-fold and reduces
its gap to attention from 2.48-fold with the reference implementations to
1.12-fold with the optimized ones (Section~\ref{subsec:short}). What remains sequential is a single
multiply--add per step, which is cheap.

A custom parallel-scan kernel therefore remains useful, but restructuring the
loop should come first: at these sequence lengths the dominant cost was ordinary elementwise work
placed inside the sequential region. The generation-time collapse of
Section~\ref{subsec:gen} is a separate problem with a separate fix---state
caching---and is not addressed by either change.

\subsection{Attention in a Hybrid Stack Retrieves Specific Tokens}
\label{subsec:localattn}

The attention-weight visualization of Section~\ref{subsec:interp} shows that,
inside the trained Hybrid model, later tokens in the example sentence attend
selectively to particular earlier content words---the pronoun ``it'' to its
referent ``tree''---rather than spreading their weight across the context or
concentrating it on immediate neighbors. Taken together with the $\alpha_t$
dynamics of Fig.~\ref{fig:alpha}, this is consistent with a division of labor
in which the Wave layers maintain a gated running summary while the Attention
layers retrieve specific earlier tokens. The observation rests on a single
example sentence (Section~\ref{subsec:limitations}).

\subsection{Limitations}\label{subsec:limitations}

Several scope limitations qualify the conclusions above. First, all models are
trained for a single epoch, at a single scale (10.94M parameters), on a single,
syntactically simple dataset \cite{eldan2023tinystories}; whether the
perplexity ranking observed here holds at larger scale, over multiple training
seeds, or on more complex, long-document corpora such as The Pile
\cite{gao2020pile} is not established. Second, the training wall-clock comparison of
Section~\ref{subsec:wallclock} and the generation-throughput comparison of
Section~\ref{subsec:gen} remain confounded by the \texttt{torch.compile}
asymmetry, since both were measured with a compiled baseline against
uncompiled recurrent configurations; only the memory and forward-latency
benchmarks are compile-free for all three. Third, no
incremental, state-caching decoding path is implemented, so the
generation-throughput results measure the current reference implementation
rather than the recurrence's ceiling. Fourth, the interpretability analysis
uses a single example sentence and is illustrative rather than systematic;
broader protocols such as HELM \cite{liang2023helm} would be a natural
extension. Fifth, the comparison includes no established sub-quadratic
baseline: benchmarking the Dynamic Wave layer against Mamba
\cite{gu2023mamba}, RWKV \cite{peng2023rwkv} and linear attention
\cite{katharopoulos2020} at a matched parameter budget is required before any
claim of relative merit within the recurrent family can be made, and a
targeted associative-recall benchmark in the style of Zoology
\cite{arora2023zoology} would test directly the recall behavior that
Section~\ref{subsec:qualitative} only illustrates. Sixth, all profiling is
performed in FP32 on a single GPU architecture; mixed-precision and
multi-device behavior may differ, particularly for the attention backends,
whose relative cost depends strongly on kernel availability.

% =====================================================================
\section{Hardware-Aware Strategy Selection Model}\label{sec:strategy}

Based on the results of Section~\ref{sec:results}, the following guidance is
proposed for choosing among the three configurations, rather than treating one
architecture as universally superior.

\begin{itemize}
\item If the best achievable language-modeling quality is the objective, use
the Hybrid Wave-Attention model: it achieves the lowest perplexity of the
three, and its peak memory is indistinguishable from the Pure Transformer, so
the quality gain is essentially free in that dimension.
\item If inference latency at short-to-moderate context is the binding
constraint, use the Pure Transformer with a fused attention kernel: it is the
fastest configuration in this regime (526.22 tokens/sec bounded-window
generation), and the optimized Wave layer only approaches, without overtaking,
its forward-pass latency.
\item Deploy the Dynamic Wave layer for its memory profile only where a fused
attention kernel is genuinely unavailable---an unusual situation in current
frameworks. Where fused attention is available, expect parity at inference and
a modest disadvantage during training.
\item Do not assume a hybrid configuration is memory-favorable by
construction, and do not assume it is unfavorable either: it tracks its most
expensive mixer, so verify against that mixer at the target context length.
\item Do not deploy the Dynamic Wave layer, in its current reference
implementation, for unbounded-context autoregressive generation: without a
state-caching decoding path its measured throughput is an order of magnitude
below the Transformer baseline.
\item Before investing in a custom parallel-scan kernel, first move all
input-dependent elementwise computation out of the sequential region; in this
study that single change gave a 2.53-fold speed-up.
\end{itemize}

% =====================================================================
\section{Conclusions}\label{sec:conclusions}

This paper formalized a Dynamic Wave recurrent layer, based on
input-dependent exponential damping and cosine phase modulation, as an
alternative and a complement to standard Multi-Head Attention, and evaluated
three 10.94M-parameter configurations under a measurement protocol that
isolates each configuration, separates mixer memory from output-layer memory,
and names the attention backend for every figure.

On quality, the results are clear and favorable: the Hybrid Wave-Attention
configuration achieves a perplexity of 5.95 on the full held-out split against
6.93 for the Transformer baseline and 7.26 for Pure Wave, and its peak memory
matches the Transformer to within 0.1\%, so this quality gain costs nothing
in that dimension.

On efficiency, the results are conditional in a way that the framing of much
of the sub-quadratic literature obscures. Pure Wave uses 78.7\% less peak VRAM
than attention at a 4096-token context when attention is computed without a
fused kernel, and exactly as much when it is; in training mode it is 8.3\%
more expensive. Similarly, hoisting elementwise terms out of the recurrence
loop sped the layer up 2.53-fold without any custom kernel, reducing its
latency gap to attention from 2.48-fold to 1.12-fold. Neither of the two
efficiency effects usually attributed to the recurrent architecture survives a
controlled measurement: the memory advantage belongs to the attention
baseline's implementation, and most of the latency penalty to the Wave layer's
own.

The broader conclusion is methodological. Efficiency claims for attention
replacements should state the attention backend, the compilation setting, the
allocator protocol, and whether the output layer is included, because in this
study such choices changed headline numbers by factors of two or more and, in
two cases, reversed the ordering they implied. The recurrence studied here is an associative scan
(Section~\ref{subsec:scan}), so the parallel form is available in principle;
what is missing is an implementation of it. Future work should supply that
kernel together with a state-caching decoding path, disentangle the
\texttt{torch.compile} asymmetry, benchmark the Dynamic Wave layer against
Mamba \cite{gu2023mamba}, RWKV \cite{peng2023rwkv} and linear attention
\cite{katharopoulos2020} at a matched budget, and extend the evaluation to
larger models, multiple seeds, targeted associative-recall tests, and longer
corpora such as The Pile \cite{gao2020pile}.

% =====================================================================
\section*{Acknowledgment}

The study was performed without financial support.

\section*{Use of Artificial Intelligence Tools}

During the preparation of this work, the authors used Claude Opus 5
(Anthropic) in order to: review the measurement code and write the benchmark
script of Sections~\ref{sec:metrics}--\ref{sec:results}; write the code that produces the figures
from the recorded results; paraphrase and format the manuscript. The Dynamic
Wave layer, the models and their training, the measurement protocol and the
conclusions are the authors' own. Every reported number was checked against
the raw measurement output, and the memory model against the measurements in
Table~\ref{tab:memmodel}. After using this tool, the authors reviewed and
edited the content as needed and take full responsibility for the content of
the publication.

% =====================================================================


\begin{thebibliography}{00}

\bibitem{radford2019} A. Radford, J. Wu, R. Child, D. Luan, D. Amodei, and
I. Sutskever, ``Language models are unsupervised multitask learners,''
\textit{OpenAI Blog}, vol. 1, no. 8, p. 9, 2019.

\bibitem{kaplan2020} J. Kaplan, S. McCandlish, T. Henighan, T. B. Brown,
B. Chess, R. Child, S. Gray, A. Radford, J. Wu, and D. Amodei, ``Scaling laws
for neural language models,'' arXiv preprint arXiv:2001.08361, 2020.

\bibitem{vaswani2017} A. Vaswani, N. Shazeer, N. Parmar, J. Uszkoreit,
L. Jones, A. N. Gomez, L. Kaiser, and I. Polosukhin, ``Attention is all you
need,'' in \textit{Proc. Advances in Neural Information Processing Systems
(NeurIPS)}, vol. 30, 2017, pp. 5998--6008.

\bibitem{dao2022} T. Dao, D. Y. Fu, S. Ermon, A. Rudra, and C. R\'e,
``FlashAttention: Fast and memory-efficient exact attention with
IO-awareness,'' in \textit{Proc. Advances in Neural Information Processing
Systems (NeurIPS)}, vol. 35, 2022, pp. 16344--16359.

\bibitem{liu2024lost} N. F. Liu, K. Lin, J. Hewitt, A. Paranjape,
M. Bevilacqua, F. Petroni, and P. Liang, ``Lost in the middle: How language
models use long contexts,'' \textit{Transactions of the Association for
Computational Linguistics}, vol. 12, pp. 157--173, 2024.

\bibitem{gu2023mamba} A. Gu and T. Dao, ``Mamba: Linear-time sequence modeling
with selective state spaces,'' in \textit{Proc. Conference on Language Modeling
(COLM)}, 2024.

\bibitem{peng2023rwkv} B. Peng et al., ``RWKV: Reinventing RNNs for the
transformer era,'' in \textit{Findings of the Association for Computational
Linguistics: EMNLP}, 2023, pp. 14048--14077.

\bibitem{de2024griffin} S. De et al., ``Griffin: Mixing gated linear
recurrences with local attention for efficient language models,'' arXiv
preprint arXiv:2402.19427, 2024.

\bibitem{lieber2024jamba} O. Lieber et al., ``Jamba: A hybrid
Transformer-Mamba language model,'' arXiv preprint arXiv:2403.19887, 2024.

\bibitem{wang2020linformer} S. Wang, B. Z. Li, M. Khabsa, H. Fang, and H. Ma,
``Linformer: Self-attention with linear complexity,'' arXiv preprint
arXiv:2006.04768, 2020.

\bibitem{choromanski2021} K. Choromanski et al., ``Rethinking attention with
performers,'' in \textit{Proc. International Conference on Learning
Representations (ICLR)}, 2021.

\bibitem{gu2022s4} A. Gu, K. Goel, and C. R\'e, ``Efficiently modeling long
sequences with structured state spaces,'' in \textit{Proc. International
Conference on Learning Representations (ICLR)}, 2022.

\bibitem{eldan2023tinystories} R. Eldan and Y. Li, ``TinyStories: How small
can language models be and still speak coherent English?'' arXiv preprint
arXiv:2305.07759, 2023.

\bibitem{dao2024flash2} T. Dao, ``FlashAttention-2: Faster attention with
better parallelism and work partitioning,'' in \textit{Proc. International
Conference on Learning Representations (ICLR)}, 2024.

\bibitem{tay2022} Y. Tay, M. Dehghani, D. Bahri, and D. Metzler, ``Efficient
transformers: A survey,'' \textit{ACM Computing Surveys}, vol. 55, no. 6,
pp. 1--28, 2022.

\bibitem{beltagy2020} I. Beltagy, M. E. Peters, and A. Cohan,
``Longformer: The long-document transformer,'' arXiv preprint
arXiv:2004.05150, 2020.

\bibitem{kitaev2020} N. Kitaev, \L. Kaiser, and A. Levskaya, ``Reformer: The
efficient transformer,'' in \textit{Proc. International Conference on Learning
Representations (ICLR)}, 2020.

\bibitem{liu2024ring} H. Liu, M. Zaharia, and P. Abbeel, ``Ring Attention with
blockwise transformers for near-infinite context,'' in \textit{Proc.
International Conference on Learning Representations (ICLR)}, 2024.

\bibitem{sun2023retnet} Y. Sun et al., ``Retentive network: A successor to
transformer for large language models,'' arXiv preprint arXiv:2307.08621,
2023.

\bibitem{katharopoulos2020} A. Katharopoulos, A. Vyas, N. Pappas, and
F. Fleuret, ``Transformers are RNNs: Fast autoregressive transformers with
linear attention,'' in \textit{Proc. International Conference on Machine
Learning (ICML)}, 2020, pp. 5156--5165.

\bibitem{gu2022s4d} A. Gu, A. Gupta, K. Goel, and C. R\'e, ``On the
parameterization and initialization of diagonal state space models,'' in
\textit{Proc. Advances in Neural Information Processing Systems (NeurIPS)},
vol. 35, 2022, pp. 35971--35983.

\bibitem{orvieto2023lru} A. Orvieto, S. L. Smith, A. Gu, A. Fernando,
C. Gulcehre, R. Pascanu, and S. De, ``Resurrecting recurrent neural networks
for long sequences,'' in \textit{Proc. International Conference on Machine
Learning (ICML)}, 2023, pp. 26670--26698.

\bibitem{poli2023hyena} M. Poli et al., ``Hyena hierarchy: Towards larger
convolutional language models,'' in \textit{Proc. International Conference on
Machine Learning (ICML)}, 2023, pp. 28043--28078.

\bibitem{fu2023h3} D. Y. Fu, T. Dao, K. K. Saab, A. W. Thomas, A. Rudra, and
C. R\'e, ``Hungry Hungry Hippos: Towards language modeling with state space
models,'' in \textit{Proc. International Conference on Learning
Representations (ICLR)}, 2023.

\bibitem{ma2023mega} X. Ma, C. Zhou, X. Kong, J. He, L. Gui, G. Neubig,
J. May, and L. Zettlemoyer, ``Mega: Moving average equipped gated attention,''
in \textit{Proc. International Conference on Learning Representations (ICLR)},
2023.

\bibitem{bahdanau2015} D. Bahdanau, K. Cho, and Y. Bengio, ``Neural machine
translation by jointly learning to align and translate,'' in \textit{Proc.
International Conference on Learning Representations (ICLR)}, 2015.

\bibitem{arora2023zoology} S. Arora, S. Eyuboglu, A. Timalsina, I. Johnson,
M. Poli, J. Zou, A. Rudra, and C. R\'e, ``Zoology: Measuring and improving
recall in efficient language models,'' in \textit{Proc. International
Conference on Learning Representations (ICLR)}, 2024.

\bibitem{blelloch1990} G. E. Blelloch, ``Prefix sums and their applications,''
School of Computer Science, Carnegie Mellon University, Tech. Rep.
CMU-CS-90-190, 1990.

\bibitem{martin2018} E. Martin and C. Cundy, ``Parallelizing linear recurrent
neural nets over sequence length,'' in \textit{Proc. International Conference
on Learning Representations (ICLR)}, 2018.

\bibitem{hendrycks2016} D. Hendrycks and K. Gimpel, ``Gaussian error linear
units (GELUs),'' arXiv preprint arXiv:1606.08415, 2016.

\bibitem{xiong2020} R. Xiong et al., ``On layer normalization in the
transformer architecture,'' in \textit{Proc. International Conference on
Machine Learning (ICML)}, 2020, pp. 10524--10533.

\bibitem{loshchilov2019} I. Loshchilov and F. Hutter, ``Decoupled weight decay
regularization,'' in \textit{Proc. International Conference on Learning
Representations (ICLR)}, 2019.

\bibitem{smith2019} L. N. Smith and N. Topin, ``Super-convergence: Very fast
training of neural networks using large learning rates,'' in
\textit{Artificial Intelligence and Machine Learning for Multi-Domain
Operations Applications}, Proc. SPIE, 2019, doi: 10.1117/12.2520589.

\bibitem{micikevicius2018} P. Micikevicius et al., ``Mixed precision
training,'' in \textit{Proc. International Conference on Learning
Representations (ICLR)}, 2018.

\bibitem{paszke2019} A. Paszke et al., ``PyTorch: An imperative style,
high-performance deep learning library,'' in \textit{Proc. Advances in Neural
Information Processing Systems (NeurIPS)}, vol. 32, 2019, pp. 8024--8035.

\bibitem{ansel2024pytorch2} J. Ansel et al., ``PyTorch 2: Faster machine
learning through dynamic graph compilation and reinforcement learning,'' in
\textit{Proc. ACM International Conference on Architectural Support for
Programming Languages and Operating Systems (ASPLOS)}, 2024, pp. 229--243.

\bibitem{tillet2019triton} P. Tillet, H. T. Kung, and D. Cox, ``Triton: An
intermediate language and compiler for tiled neural network computations,'' in
\textit{Proc. 3rd ACM SIGPLAN International Workshop on Machine Learning and
Programming Languages (MAPL)}, 2019, pp. 10--19.

\bibitem{gao2020pile} L. Gao et al., ``The Pile: An 800GB dataset of diverse
text for language modeling,'' arXiv preprint arXiv:2101.00027, 2020.

\bibitem{shah2024flash3} J. Shah, G. Bikshandi, Y. Zhang, V. Thakkar,
P. Ramani, and T. Dao, ``FlashAttention-3: Fast and accurate attention with
asynchrony and low-precision,'' in \textit{Proc. Advances in Neural
Information Processing Systems (NeurIPS)}, vol. 37, 2024.

\bibitem{dao2024mamba2} T. Dao and A. Gu, ``Transformers are SSMs: Generalized
models and efficient algorithms through structured state space duality,'' in
\textit{Proc. International Conference on Machine Learning (ICML)}, 2024.

\bibitem{yang2025gdn} S. Yang, J. Kautz, and A. Hatamizadeh, ``Gated delta
networks: Improving Mamba2 with delta rule,'' in \textit{Proc. International
Conference on Learning Representations (ICLR)}, 2025.

\bibitem{waleffe2024} R. Waleffe et al., ``An empirical study of Mamba-based
language models,'' arXiv preprint arXiv:2406.07887, 2024.

\bibitem{minimax2025} MiniMax et al., ``MiniMax-01: Scaling foundation models
with lightning attention,'' arXiv preprint arXiv:2501.08313, 2025.

\bibitem{liang2023helm} P. Liang et al., ``Holistic evaluation of language
models,'' \textit{Transactions on Machine Learning Research}, 2023.

\end{thebibliography}
\end{document}